\documentclass{journal}

\usepackage{hyperref}
\usepackage{cleveref}

\usepackage{graphicx}
\usepackage{epstopdf}
\usepackage{epsfig}
\usepackage{cite,footnote,xspace,syntonly,algorithm,algorithmic,bm}

\usepackage{rotating}

\usepackage[utf8]{inputenc} 
\usepackage[T1]{fontenc}    
\usepackage{hyperref}       
\usepackage{url}            
\usepackage{booktabs}       
\usepackage{amsfonts}       
\usepackage{microtype}      
\usepackage{amsmath}
\usepackage{bbm}
\usepackage{bm}
\usepackage{amssymb}

\usepackage[final]{nips_2017}

\def \cN {\mathcal{N}}

\def \cM {\mathcal{M}}
\def \cX {\mathcal{X}}

\def \cD {\mathcal{D}}
\def \cG {\mathcal{G}}
\def \cF {\mathcal{F}}

\def \bPhi {\boldsymbol{\Phi}}
\def \bTheta {\boldsymbol{\Theta}}

\def \by {\mathbf{y}}
\def \bx {\mathbf{x}}

\def \be {\mathbf{e}}
\def \bv {\mathbf{v}}

\def \bI {\mathbf{I}}

\def \bX {\mathbf{X}}

\title{Deep Generative Adversarial Networks for Compressed Sensing (GANCS) Automates MRI}

\author{\\{\it \large{Morteza Mardani$^{1,3}$, Enhao Gong$^{1}$, Joseph Y. Cheng$^{1,2}$, Shreyas Vasanawala$^{2}$,}} \\ \it{ \large{Greg Zaharchuk$^{2}$, Marcus  Alley$^{2}$, Neil Thakur$^{2}$, Song Han$^{4}$, William Dally$^{4}$,}} \\ {\it \large{John M. Pauly$^{1}$, and Lei Xing$^{1,3}$}} \thanks{The authors are with the Stanford University, Departments of Electrical Engineering$^{1}$, Radiology$^{2}$, Radiation Oncology$^{3}$, and Computer Science$^{4}$. } \\ }

\begin{document}

\maketitle 

\begin{abstract}
Magnetic resonance image (MRI) reconstruction is a severely ill-posed linear inverse task demanding time and resource intensive computations that can substantially trade off {\it accuracy} for {\it speed} in real-time imaging. In addition, state-of-the-art compressed sensing (CS) analytics are not cognizant of the image {\it diagnostic quality}. To cope with these challenges we put forth a novel CS framework that permeates benefits from generative adversarial networks (GAN) to train a (low-dimensional) manifold of diagnostic-quality MR images from historical patients. Leveraging a mixture of least-squares (LS) GANs and pixel-wise $\ell_1$ cost, a deep residual network with skip connections is trained as the generator that learns to remove the {\it aliasing} artifacts by projecting onto the manifold. LSGAN learns the texture details, while $\ell_1$ controls the high-frequency noise. A multilayer convolutional neural network is then jointly trained based on diagnostic quality images to discriminate the projection quality. The test phase performs feed-forward propagation over the generator network that demands a very low computational overhead. Extensive evaluations are performed on a large contrast-enhanced MR dataset of pediatric patients. In particular, images rated based on expert radiologists corroborate that GANCS retrieves high contrast images with detailed texture relative to conventional CS, and pixel-wise schemes. In addition, it offers reconstruction under a few milliseconds, two orders of magnitude faster than state-of-the-art CS-MRI schemes. 
\end{abstract}

\section{Introduction}
Owing to its superb soft tissue contrast, magnetic resonance imaging (MRI) nowadays serves as the major imaging modality in clinical practice. Real-time MRI visualization is of paramount importance for diagnostic and therapeutic guidance for instance in next generation platforms for MR-guided, minimally invasive neurosurgery~\cite{clearpoint}. However, the scan is quite slow, taking several minutes to acquire clinically acceptable images. This becomes more pronounced for high-resolution and volumetric images. As a result, the acquisition typically undergoes significant undersampling leading reconstruction to a seriously ill-posed linear inverse problem. To render it well-posed, the conventional compressed-sensing (CS) incorporates the prior image information by means of sparsity regularization in a proper transform domain such as Wavelet (WV), or, Total Variation (TV); see e.g.,~\cite{pualy_mri20017}. This however demands running iterative optimization algorithms that are time and resource intensive. This in turn hinders {\it real-time} MRI visualization and analysis.

Recently, a few attempts have been carried out to {\it automate} medical image reconstruction by leveraging historical patient data; see e.g.,~\cite{Majumdar'15,lowdose_ct2017}. They train a network that maps the aliased image to the gold-standard one using convolutional neural networks (CNN) with residuals for computed tomography (CT)~\cite{lowdose_ct2017}, denoising auto-encoders for MRI~\cite{Majumdar'15}. Albeit, speed up, they suffer from blurry and aliasing artifacts. This is mainly due to adopting a pixel-wise $\ell_1$/$\ell_2$ cost that is oblivious of high-frequency texture details, which is crucial for drawing diagnostic decisions. See also the recent DeepADMM scheme in \cite{deepADMM2016} for CS MRI that improves the quality, but it is as slow as the conventional CS. Generative adversarial networks (GANs) have been lately proved very successful in~\cite{gan-goodfellow2014,dcgan2016} modeling a low-dimensional distribution (manifold) of natural images that are perceptually appealing~\cite{Zhu et al'16}. In particular, for image super-resolution tasks GANs achieve state-of-the-art perceptual quality under $4\times$ upscaling factor for natural images e.g., from ImageNet~\cite{leding et al'16, Sonderby et al'14}. GANs has also been deployed for image inpaitning~\cite{inpainting-yeh-2016}, style transfer~\cite{johnson2016}, and visual manipulation~\cite{Zhu et al'16}.

Despite the success of GANs for {\it local} image restoration such as super-resolution and inpainting, to date, they have not been studied for removing {\it aliasing} artifacts in biomedical image reconstruction tasks. This is indeed a more difficult image restoration tasks. In essence, aliasing artifacts (e.g., in MRI) emanate from data undersampling in a different domain (e.g., Fourier, projections) which {\it globally} impact image pixels. Inspired by the high texture quality offered by GANs, and the high contrast of MR images, we employ GANs to learn a low-dimensional manifold of diagnostic-quality MR images. To this end, we train a tandem network of a generator (G) and a discriminator (D), where the generator aims to generate the ground-truth images from the complex-valued aliased ones using a deep residual network (ResNet) with skip connections, with refinement to ensure it is consistent with measurement (data consistency). The aliased input image is simply obtained via inverse Fourier Transform (FT) of undersampled data. D network then scores the G output, using a multilayer convolutional neural network (CNN) that scores one if the image is of diagnostic quality, and, zero if it contains artifacts. For training we adopt a mixture of LSGAN~\cite{lsgan2017} and $\ell_1$ pixel-wise criterion to retrieve high-frequency texture while controlling the noise. We performed evaluations on a large cohort of pediatric patients with contrast-enhanced abdominal images. The retrieved images are rated by expert radiologists for diagnostic quality. Our observations indicate that GANCS results have almost similar quality to the gold-standard fully-sampled images, and are superior in terms of diagnostic quality relative to the existing alternatives including conventional CS (e.g., TV and WV), $\ell_2$-, and $\ell_1$-based criteria. Moreover, the reconstruction only takes around $10-20$ msec, that is two orders of magnitude faster than state-of-the-art conventional CS toolboxes.

Last but not least, the advocated GANCS scheme tailors inverse imaging tasks appearing in a wide range of applications with budgeted acquisition and reconstruction speed. All in all, relative to the past work this paper's main contributions are summarized as follows:

\begin{itemize}

	\item Propose GANCS as a data-driven regularization scheme for solving ill-posed linear inverse problems that appear in imaging tasks dealing with (global) aliasing artifacts 
	
	\item First work to apply GAN as a automated (non-iterative) technique for aliasing artifact suppression in MRI with state-of-the-art image diagnostic quality and reconstruction speed

	\item Proposed and evaluated a novel network architecture to achieve better trade-offs between data-consistency (affine projection) and manifold learning 
			
	\item Extensive evaluations on a large contrast-enhanced MRI dataset of pediatric patients, with the reconstructed images rated by expert radiologists

\end{itemize}

The rest of this paper is organized as follows. Section 2 states the problem. Manifold learning using LSGANs is proposed in Section 3. Section 4 also reports the data evaluations, while the conclusions are drawn in Section 5.

\section{Problem Statement}
\label{sec:problem_statement}
Consider an ill-posed linear system $\by=\bPhi \bx + \bv$ with $\bPhi \in \mathbb{C}^{M \times N}$ where $M \ll N$, and $\bv$ captures the noise and unmodeled dynamics. Suppose the unknown and complex-valued image $\bx$ lies in a {\it low-dimensional} manifold, say $\cM$. No information is known about the manifold besides the training samples $\cX:=\{\bx_k\}_{k=1}^K$ drawn from it, and the corresponding (possibly) noisy observations $\mathcal{Y}:=\{\by_k\}_{k=1}^K$. Given a new observation $\by$, the goal is to recover $\bx$. For instance, in the MRI context motivated for this paper $\bPhi$ refers to the partial 2D FT that results in undersampled $k$-space data $\by$. To retrieve the image, in the first step we learn the manifold $\cM$. Subsequently, the second step projects the aliased image, obtained via e.g., pseudo inverse $\bPhi^{\dagger}\by$ onto $\cM$ to discard the artifacts. For the sake of generality, the ensuing is presented for a generic linear map $\bPhi$.

\section{Manifold Learning via Generative Adversarial Networks}
\label{sec:gans}
The inverse imaging solution is to find solutions of the intersection between two subspaces defined by acquisition model and image manifold. In order to effectively learn the image manifold from the available (limited number of) training samples we first need to address the following important questions:

\begin{itemize}

\item How to ensure the trained manifold contains plausible images? 

\item How to ensure the points on the manifold are data consistent, namely $\by \approx \bPhi\bx, ~\forall \bx \in \cM$?



\end{itemize}
To address the first question we adopt GANs, that have recently proven very successful in estimating prior distribution for images. GANs provide sharp images that are visually plausible~\cite{gan-goodfellow2014}. In contrast, variational autoencoders~\cite{leding et al'16}, a important class of generative models, use pixel-wise MSE costs that results in high pick signal-to-noise ratios but often produce overly-smooth images that have poor perceptual quality. Standard GAN consists of a tandem network of G and D networks. Consider the undersampled image $\tilde{\bx}:=\bPhi^{\dagger}\by$ as the input to the G network. The G network then projects $\tilde{\bx}$ onto the low-dimensional manifold $\cM$ containing the high-quality images $\cX$. Let $\hat{\bx}$ denote the output of G, it then passes through the discriminator network D, that outputs one if $\hat{\bx} \in \cX$, and zero otherwise.

\begin{figure}[t]
	\centering
	\hspace{-0.25cm}\includegraphics[scale=0.925]{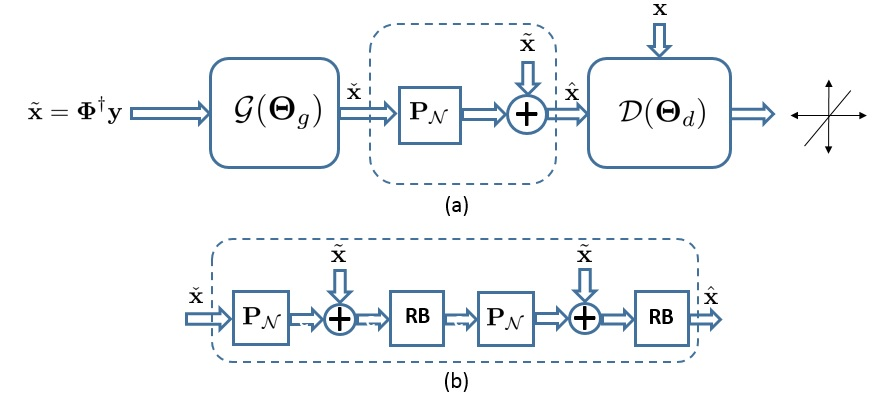}
	\caption{(a) GANCS structure for manifold learning, where the dashed module is projection on the feasible set. (b) The multilayer residual blocks (RB) for data consistency. }
	\label{fig:fig_net}
\end{figure}

The output of G, namely $\check{\bx}$, however may not be consistent with the data. To tackle this issue, we add another layer after G that projects onto the feasible set of $\by=\bPhi\bx$ to arrive at $\hat{\bx}=\bPhi^{\dagger} \by + (\bI-\bPhi^{\dagger}\bPhi) \check{\bx}$. Alternatively, we can add a soft LS penalty when training the G network, as will be seen later in (P1). To further ensure that $\hat{\bx}$ lies in the intersection of the manifold $\cM$ and the space of data consistent images we can use a mutlilayer network that alternates between residual units and data consistency projection as depicted in Fig.~\ref{fig:fig_net} (b). We have observed that using only a couple of residual units may improve the performance of G in discarding the aliasing artifacts. The overall network architecture is depicted in Fig.~\ref{fig:fig_net} (a), where $\mathbf{P}_{\cN}:=(\bI-\bPhi^{\dagger}\bPhi)$ signifies projection onto the nullspace of $\bPhi$.



Training the network in Fig.~\ref{fig:fig_net} amounts to playing a game with conflicting objectives between the adversary G and the discriminator D. D network aims to score one the real images drawn from the data distribution $p_{x}$, and score zero the rest. G network also aims to map the input images $\tilde{\bx}=\bPhi^{\dagger}\by$ with the distribution $p_{\tilde{x}}=p_{x}(\bPhi^{\dagger}\bPhi \bx)$ to the fake images $\hat{\bx}$ that fool the D network. Various strategies have been devised to reach the equilibrium. They mostly differ in terms of the cost function adopted for the G and D networks~\cite{gan-goodfellow2014}, \cite{lsgan2017}. The standard GAN uses a sigmoid cross-entropy loss that leads to vanishing gradients which renders the training unstable, and as a result it suffers from sever degrees of mode collapse. In addition, for the generated images classified as the real with high confidence (i.e., large decision variable), no cost is incurred. Hence, the standard GAN tends to pull samples away from the decision boundary, that introduces non-realistic images~\cite{lsgan2017}. LSGN instead pulls the generated samples towards the decision boundary by using a LS cost. 

One issue with GAN however is that it introduces high frequency noise all over the image. $\ell_1$ criterion has proven well in discarding the noise from natural images as it does appropriately penalize the low-intensity noise~\cite{lossfunction_zhao2017}. Accordingly, to reveal fine texture details while discarding noise, we are motivated to adopt a mixture of LSGAN and $\ell_1$ costs to train the generator. The overall procedure aims to jointly minimize the discriminator cost
\begin{align}
{\rm(P1.1)} \quad \quad \min_{\bTheta_d}~ \mathbb{E}_{\bx} \Big[\Big(1-\cD(\bx;\bTheta_d)\Big)^2\Big] +  \mathbb{E}_{\by} \Big[\Big(\cD(\cG(\bPhi^{\dagger} \by; \bTheta_g);\bTheta_d)\Big)^2\Big]  \nonumber
\end{align}
and the generator cost
\begin{align}
{\rm(P1.2)} \quad \quad \min_{\bTheta_g}~ \mathbb{E}_{\by}\Big[\Big\|\by - \bPhi \cG(\bPhi^{\dagger} \by; \bTheta_g) \Big\|^2 \Big] & + \eta \mathbb{E}_{\bx,\by}\Big[\Big\|\bx - \cG(\bPhi^{\dagger} \by; \bTheta_g) \Big\|_1 \Big]  \nonumber \\ & + \lambda \mathbb{E}_{\by} \Big[\Big(1-\cD \big(\cG(\bPhi^{\dagger} \by; \bTheta_g);\bTheta_d\big)\Big)^2\Big] \nonumber
\end{align}
The first LS fitting term in (P1.2) is a soft penalty to ensure the input to D network is data consistent. Parameters $\lambda$ and $\eta$ also control the balance between manifold projection, noise suppression and data consistency.

Looking carefully into (P1.2) the generator reconstructs image $\cG(\bPhi^{\dagger} \by; \bTheta_g)$ from the data $\by$ using an expected regularized-LS estimator, where the regularization is learned form training data via LSGAN and $\ell_1$-net. Different from the conventional CS formulation which also optimize the reconstruction with $\ell_1$-regularized LS estimation, the entire optimization only happens in training and the generator learned can be directly applied to new samples to achieve fast reconstruction.

As argued in~\cite{lsgan2017}, it can be shown that LSGAN game yields minimizing the Pearson-$\chi^2$ divergence. For (P1) following the same arguments as of the standard GANS in~\cite{gan-goodfellow2014} and~\cite{lsgan2017} it can be readily shown that even in the presence of LS data consistency and $\ell_1$ penalty, the distribution modeled by G network, say $p_g$, coincides with the true data distribution. This is formally stated next.

\noindent\textbf{Lemma 1.}~{\it For the noise-free scenario ($\bv=\mathbf{0}$), suppose D and G have infinite capacity. Then, for a given generator network G, i) the optimal discriminator D is $\cD^{*}(\bTheta_d;\check{\bx})=p_x(\check{\bx})/(p_x(\check{\bx})+p_g(\check{\bx}))$; and ii) $p_g=p_x$ achieves the equilibrium for the game (P1). }

\noindent{\it Proof.} The first part is similar to the one in~\cite{lsgan2017} with the same cost for D. The second part also readily follows as the LS data consistency and $\ell_1$ penalty are non-negative, and become zero when $p_g=p_x$. Thus, according to Pearson-$\chi^2$ divergence still bounds (P1.2) objective from below, and is achievable when $p_g=p_x$. $\blacksquare$

%
%
%
%
%

\subsection{Stochastic alternating minimization}
\label{subsec:alt_min}
To train the G and D networks, a mini-batch stochastic alternating minimization scheme is adopted. At $k$-th iteration with the mini-batch training data $\{(\bx_{\ell},\by_{\ell})\}_{\ell=1}^L$, assuming that G is fixed, we first update the discriminator $\bTheta_d$ by taking a single descent step with momentum along the gradient of D cost, say $f_d$. Similarly, given the updated $\bTheta_d$, the G network is updated by taking a gradient descent step with momentum along the gradient of G cost, say $f_g$. The resulting iterations are listed under Algorithm~\ref{tab:alg_gan_training}, where the gradients $\nabla_{\bTheta_g}\cG(\bTheta_g;\tilde{\bx}_{\ell})$, $\nabla_{\bTheta_d}\cD(\hat{\bx}_{\ell};\bTheta_d)$, and $\nabla_{\bTheta_g}\cD(\cG(\bTheta_g;\tilde{\bx}_{\ell});\bTheta_d)$ are readily obtained via backpropagation over D and G networks. Also, $G_n$ refers to the $n$-th output pixel of G network, and $[.]_n$ picks the $n$-th pixel.  

%



\begin{algorithm}[t]
	\caption{Training algorithm using BP based stochastic alternating minimization} \small{
		\begin{algorithmic}
			\STATE \textbf{input} 
			$\{(\bx_{\ell},\by_{\ell})\}_{\ell=1}^{L},\lambda,\mu, \bPhi$.
			\STATE \textbf{initialize} $(\bTheta_g[0],\bTheta_d[0])$ at random.
			
			\FOR {${\rm epoch}=1,\ldots,{\rm epoch}_{\max}$}
			
			\FOR {$k=1,\ldots,L/L_b$}
			
			\STATE \textbf{S1) Random mini-batch selection} 
			 
			\STATE {\rm Sample the mini-batch $\{\by_{\ell}\}_{\ell=1}^{L_b}$, and define $\tilde{\bx}:=\bPhi^{\dagger} \by_{\ell}$, $\hat{\bx}:=\cG(\bTheta_g;\tilde{\bx}_{\ell})$, and $\hat{\be}_{\ell} := \by_{\ell} - \bPhi \hat{\bx}$}
			
		    \STATE \textbf{S2) Discriminator update:} gradient-descent with momentum along  
			
            \STATE $ \nabla_{\bTheta_d} C_d := \frac{ \mu}{L_b} \sum_{\ell=1}^{L_b} \Big\{ -(1-\cD(\bx_{\ell};\bTheta_d)) \nabla_{\bTheta_d} \cD(\bx_{\ell};\bTheta_d) + \cD(\hat{\bx}_{\ell};\bTheta_d) \nabla_{\bTheta_d} \cD(\hat{\bx}_{\ell};\bTheta_d) \Big\}$
            
            
            \STATE \textbf{S3) Generator update:} gradient-descent with momentum along  
         		    

            \STATE $\nabla_{\bTheta_g} C_g := \frac{\mu}{L_b} \sum_{\ell=1}^{L_b} \sum_{n=1}^N \Big\{  -\lambda (1-\cD(\hat{\bx}_{\ell};\bTheta_d)) [\nabla_{\hat{\bx}} \cD(\hat{\bx}_{\ell};\bTheta_d)]_n -  \boldsymbol{\phi}_n^{\top} \hat{\be}_{\ell} + \eta [\rm{sgn}(\bx_{\ell} - \cG(\bTheta_g;\tilde{\bx}) )]_n \Big\} \nabla_{\bTheta_g} \cG_n(\bTheta_g;\tilde{\bx})$

			
			\ENDFOR
			
			\ENDFOR
			
			\RETURN  $(\bTheta_g, \bTheta_d)$
						
		\end{algorithmic}}
		\label{tab:alg_gan_training}
	\end{algorithm}



\section{Experiments}
\label{sec:eval}
Effectiveness of the novel GANCS scheme is assessed in this section via tests for MRI reconstruction. A single-coil MR acquisition model is considered where for $n$-th patient the acquired $k$-space data abides to $y_{i,j}^{(n)} = [\cF(\bX_n)]_{i,j} + v_{i,j}^{(n)},~~(i,j) \in \Omega$. Here, $\cF$ is the 2D FT, and the set $\Omega$ indexes the sampled Fourier coefficients. As it is conventionally performed with CS MRI, we select $\Omega$ based on a variable density sampling with radial view ordering~\cite{} that tends to pick low frequency components from the center of $k$-space (see sampling mask in Fig. 4 (left) of the supplementary document). Throughout the test we assume $\Omega$ collects only $20\%$ of the Fourier coefficients, and we choose $\lambda=0.1$. 
%




\noindent{\textbf{Dataset}.}~High contrast abdominal image volumes are acquired for $350$ pediatric patients after gadolinium-based contrast enhancement. Each 3D volume includes contains $151$ axial slices of size $200 \times 100$. Axial slices used as input images for training a neural network. $300$ patients ($45,300$ images) are considered for training, and $50$ patients ($7,550$ images) for test. All in vivo scans were acquired at the Stanford’s Lucile Packard Children’s Hospital on a 3T MRI scanner (GE MR750) with voxel resolution $1.07 \times 1.12 \times 2.4$ mm.

Under this setting, the ensuing parts address the following questions:

Q1. How does the perceptual cost learned by GANCS improve the image quality compared with the pixel-wise $\ell_2$ and $\ell_1$ costs? 

Q2. How much speed up and quality improvement one can achieve using GANCS relative to conventional CS? 

Q3. What MR image features derive the network to learn the manifold and remove the aliasing artifacts?

Q4. How many samples/patients are needed to achieve a reasonable image quality?



\subsection{Training and network architecture}
\label{subsec:training}
The input and output are complex-valued images of the same size and each include two channels for real and imaginary components. The input image $\tilde{\bx}$ is simply generated using inverse 2D FT of the sampled $k$-space, which is severely contaminated by artifacts. Input channels are then convolved with different kernels and added up in the next layer. Note, all network kernels are assumed real-valued. Inspired by super-resolution ideas in~\cite{johnson2016, leding et al'16}, and the network architecture in ~\cite{srez} we adopt a deep residual network for the generator with $8$ residual blocks. Each block consists of two convolutional layers with small $3 \times 3$ kernels and $64$ feature maps that are followed by batch normalization and ReLU activation. It then follows by three convolutional layers with map size $1 \times 1$, where the first two layers undergo ReLU activation, while the last layer has sigmoid activation to return the output. G network learns the projection onto the manifold while ensuring the data consistency at the same time, where the manifold dimension is controlled by the number of residual blocks and feature maps and the settings of discriminator D network.


To satisfy data consistency term, previous work in the context of image super-resolution~\cite{Sonderby et al'14} used (hard) affine projection after the G network. However, the affine projection drifts $\hat{\bx}$ away from the manifold landscape. As argued in Section 3, we instead use a multilayer succession of affine projection and convolutional residual units that project back $\hat{\bx}$ onto the manifold. We can repeat this procedure a few times to ensure $\hat{\bx}$ lies close to the intersection. This amounts to a soft yet flexible data consistency penalty.

The D network starts from the output of the G network with two channels. It is composed of $8$ convolutional layers. In all the layers except the last one, the convolution is followed by batch normalization, and subsequently ReLU activation. No pooling is used. For the first four layers, number of feature maps is doubled from $8$ to $64$, while at the same time convolution with stride $2$ is used to reduce the image resolution. Kernel size $3 \times 3$ is adopted for the first 5 layers, while the last two layers use kernel size $1 \times 1$. In the last layer, the convolution output is averaged out to form the decision variable for binary classification. No soft-max is used.

Adam optimizer is used with the momentum parameter $\beta=0.9$, mini-batch size $L_b=8$, and initial learning rate $\mu=10^{-5}$ that is halved every $5,000$ iterations. Training is performed with TensorFlow interface on a NVIDIA Titan X Pascal GPU, 12GB RAM. We allow $20$ epochs that takes around $6$ hours for training. The implementation is available online at~\cite{github-gancs-2017}. 


As a figure of merit for image quality assessment we adopt SNR (dB), and SSIM that is defined on a cropped window of size $50 \times 50$ from the center of axial slices. In addition, we asked Radiologists Opinion Score (ROS) regarding the diagnostic quality of images. ROS ranges from $1$ (worse) to $5$ (excellent) based on the overall images quality in terms of sharpness/blurriness, and appearance of residual artifacts. 


\subsection{Observations and discussion}
\label{subsec:obs}
\begin{figure}[t]
	\hspace{-1.85cm}\includegraphics[scale=0.66]{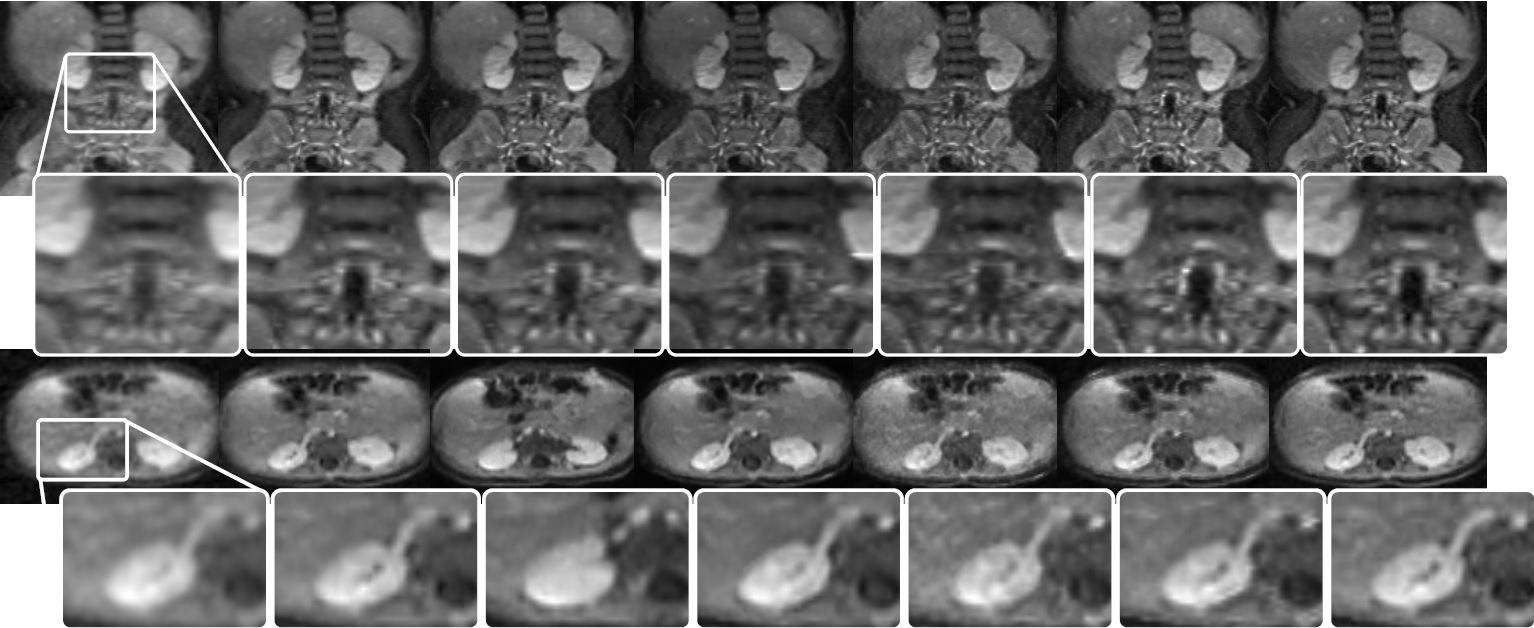}
	\caption{Representative coronal (1st row) and axial (3rd row) images for a test patient retrieved by ZF (1st), CS-WV (2nd), $\ell_2$-net (3th), $\ell_1$-net (4th), GAN (5th), GANCS (6th), and gold-standard (7th). }
	\label{fig:fig_recon_gancs_vs_cs_mse_4fold}
\end{figure}

Retrieved images by various methods are depicted in Fig.~\ref{fig:fig_recon_gancs_vs_cs_mse_4fold} with $5$-fold undersampling of $k$-space. For a random test patient, representative slices from axial, and coronal orientations, respectively, are shown from top to bottom. Columns from left to right also show, respectively, the images reconstructed by zero-filling (ZF), CS-WV, CS-TV, $\ell_2$-net, $\ell_1$-net, GAN, GANCS with $\lambda=\eta=10$, and the gold-standard (GS). Note, we propose $\ell_1$-net and $\ell_2$-net using the same network structure and training as in Section~\ref{subsec:training}, with only changing the G net cost function in (P1). CS reconstruction is performed using the Berkeley Advanced Reconstruction Toolbox (BART)~\cite{bart2016}, where the tunning parameters are optimized for the best performance. GANCS, $\ell_1$-net and $\ell_2$-net are trained with ZF images that apparently contain aliasing artifacts.

Quantitative metrics including the SNR (dB), SSIM, and the reconstruction time (sec) are also reported in Table I. These metrics are averaged out over all axial slices for test patients. As apparent from the magnified regions, GANCS returns the most detailed images with high contrast and texture details that can reveal the small structures. $\ell_2$-net images are seen somehow over-smoothed as the $\ell_2$ cost encourages finding pixel-wise averages of plausible solutions. Also, $\ell_1$-net performs better than $\ell_2$-net, which was also already reported in a different setting~\cite{lossfunction_zhao2017}, but still not as sharp as GANCS which leverages both $\ell_1$-net and GAN. GAN results with $\eta=0$ also introduces sharp images but noise is still present all over the image. CS-based results are also depicted as the benchmark MR reconstruction scheme nowadays, where evidently introduce blurring artifacts.

CS-based scheme achieve higher SNR and SSIM, but they miss the high frequency textures as evidenced by Fig.~\ref{fig:fig_recon_gancs_vs_cs_mse_4fold}. In addition, they demands iterative algorithms for solving non-smooth optimization programs that takes a few seconds for reconstruction using the optimized BART toolbox~\cite{bart2016}. In contrast, the elapsed time for GANCS is only about $10$ msec, which allows reconstructing $100$ frames per second, and thus a suitable choice for real-time MRI visualization tasks. Regarding the convergence, we empirically observe faster and more stable training by imposing more weight on the data consistency which restricts the search space for the network weights.

\begin{figure}[t]
	\centering
	\begin{tabular}{c}
		\hspace{-8mm}\epsfig{file=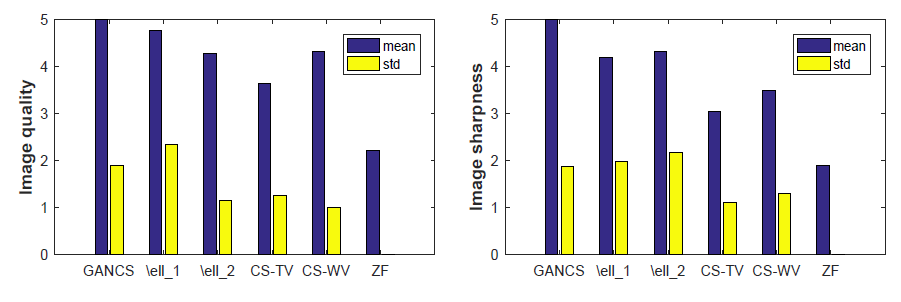,width=1.1
			\linewidth, height=1.95 in }  \\

	\end{tabular}
	\caption{Mean and standard deviation of image quality artifacts and blurriness scored by expert radiologists for various reconstruction techniques. Scores $1$ to $5$ rate from poor to excellent. }
	\label{fig:fig_barplot_ros}
\end{figure}

To assess the perceptual quality of resulting images we also asked the opinion of expert radiologists. We normalize the scores so as the gold-standard images are rated excellent (i.e., ROS=$5$). Statistical ROS is evaluated for the image quality, residual artifacts, and image sharpness. It is shown in the bar plot of Fig.~\ref{fig:fig_barplot_ros}, which confirms GANCS almost perceptually pleasing as the gold-standard scan. This demonstrates the superior diagnostic quality of GANCS images relative to the other alternatives.

For the sake of completeness, the evolution of different (empirical) costs associated with the generator cost in (P1.2) over batches are also depicted in Fig.~\ref{fig:fig_loss}. It is observed that the data consistency cost and GAN loss tend to improve alternatively to find the distribution at the intersection of manifold and dats consistency space. 




%
%
%
%


\begin{table}[t]
	\caption{Average SNR (dB), SSIM, ROS, and reconstruction time (sec) comparison of different schemes under $5$-fold undersampling.  }
	\vspace{-2.5mm}
	\label{tab:table_comp_quantitative}
	\begin{center}
		\begin{tabular} {|c|c|c|c|c|c|c|c|}
			\hline
			Scheme &  ZF & CS-WV & CS-TV &  $\ell_2$-net & $\ell_1$-net & GAN  & GANCS  \\
			\hline\hline
			SNR & $15.28$ & $20.74$ & $21.33$ & $18.96$ & $18.64$ & $16.6$ & $20.48$     \\
			\hline
			SSIM & $0.72$ & $0.88$ & $0.87$ & $0.81$  & $0.79$ & $0.78$ & $0.87$    \\
			\hline
			Recon. time & $5\hspace{-1mm} \times \hspace{-1mm} 10^{-4}$ & $5.27$ & $1.51$  & $0.02$ & $0.02$ & $0.02$ & $0.02$  \\
			\hline
			
		\end{tabular}
	\end{center}
\end{table}

\begin{figure}[t]
	\centering
	\hspace{0.0cm}\includegraphics[scale=1]{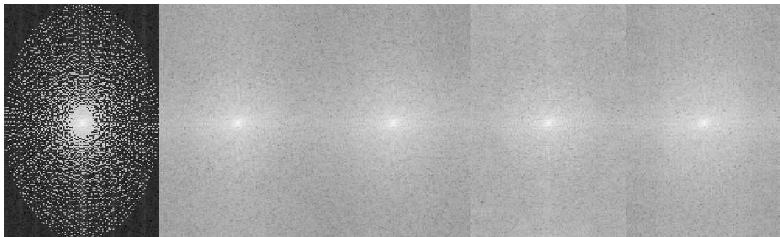}
	\caption{Representaitve $k$-space axial image retrieved by ZF (1st column), CS-WV (2nd), CS-TV (3rd), and GANCS (4th), and gold-standard (5th). }
	\label{fig:fig_recon_kspace_gancs_vs_cs_mse_4fold}
\end{figure}

\begin{figure}[t]
	\centering
	\hspace{0cm}\includegraphics[scale=0.65]{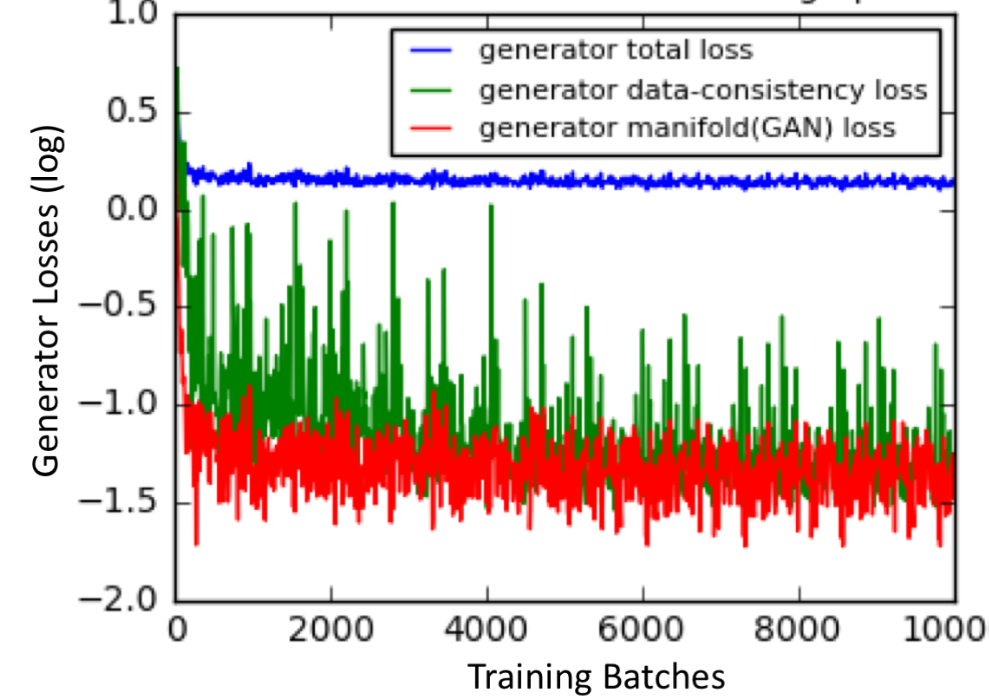}
	\caption{Evolution of different costs contributing in the overall training cost of G network.  }
	\label{fig:fig_loss}
\end{figure}





\noindent\textbf{Manifold landscape.}~We visualize what the discriminator learns by showing the feature maps in different layers as heat-maps superimposed on the original images. Since there are several feature maps per layer, we computed the Principle Component maps for each layer and visualize the first $8$ dominant ones. Fig.~ indicates that after learning from tens of thousands of generated MRI images by the G network and their gold standards including different organs, is able to detect anatomically valuable features. It is observed that the first layers reveal the edges, while the last layers closer to the classification output reveal more regions of interests that include both anatomy and texture details. This observation is consistent with the way expert radiologist inspect the images based on their diagnosis quality.

\begin{figure}[ht!]
	\centering
	\hspace{0cm}\includegraphics[scale=0.17]{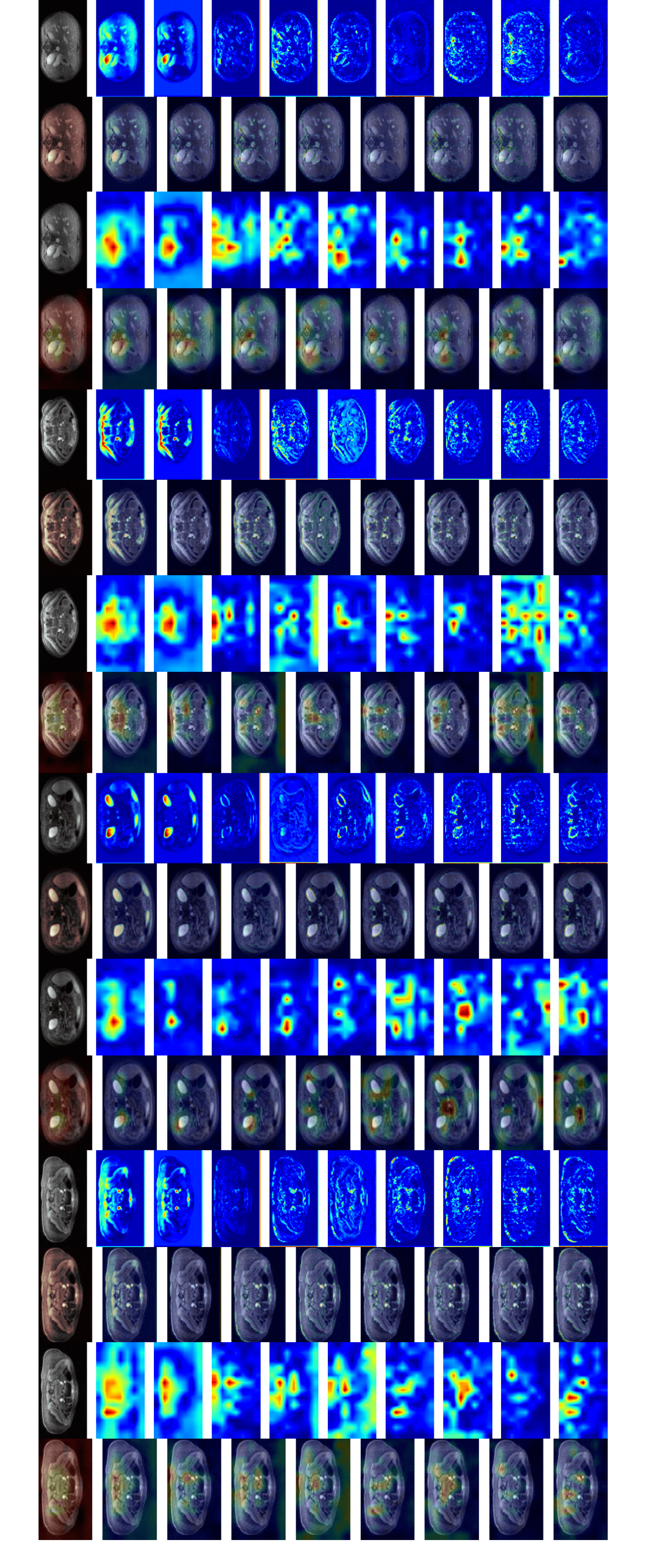}
	\caption{Heat-map of discriminator feature maps at four layers for four different images. Each 4 row from top to bottom represent the results from one MR image. The first row shows the MR image and the Principle Components of the network features from the first layer. The second row shows an overlay view of the MR image and the heat-map. The third row shows the MR slice image with the Principle Components of the network features from the last layer of discriminator; while the fourth row shows the overlay view of the MR image and the heat-maps.}
	\label{fig:fig_heatmap}
\end{figure}

\begin{figure}[t]
	\centering
	\hspace{0.0cm}\includegraphics[scale=0.15]{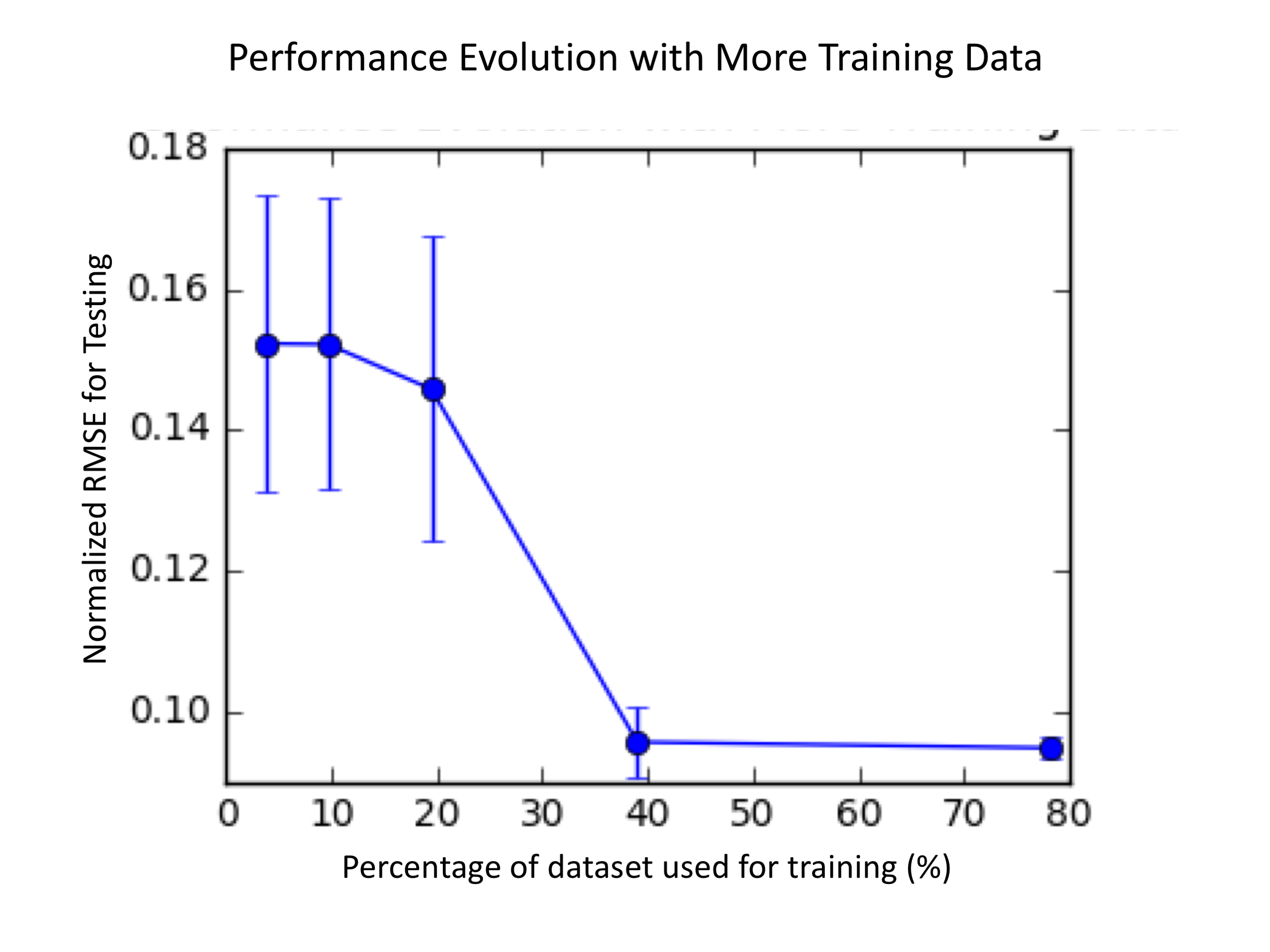}
	\caption{Performance changes with different size of dataset used for training (output about 45,300 images) }
	\label{fig:fig_performance}
\end{figure}

\noindent\textbf{Performance with different number of patients}~We also experimented on the number of patients needed for training and achieving good reconstruction quality in the test phase. It is generally valuable for the clinicians how much training data is needed as in the medical applications, patient data is not easily accessible due to privacy concerns. Fig.~\ref{fig:fig_performance} plots the normalized RMSE on a test set versus the percentage of patients used for training (normalized by the maximum patient number $350$). Note, the variance differences for different training may be due to the training with fewer samples has better convergence, since we are using the same epoch numbers for all the training cases. More detailed study is the subject of our ongoing research.

\section{Conclusions and Future Work}
\label{sec:conclusion}
This paper caters a novel CS framework that leverages the historical data for faster and more diagnosis-valuable image reconstruction from highly undersampled observations. A low-dimensional manifold is learned where the images are not only sharp and high contrast, but also consistent with both the real MRI data and the acquisition model. To this end, a neural network based on LSGANs is trained that consists of a generator network to map a readily obtainable undersmapled image to the gold-standard one. Experiments based on a large cohort of abdominal MR data, and the evaluations performed by expert radiologists confirm that the GANCS retrieves images with better diagnostic quality in a real-time manner (about $10$ msec, more than $100$ times faster than state-of-the-art CS MRI toolbox). This achieves a significant speed-up and diagnostic accuracy relative to standard CS MRI. Last but not least, the scope of the novel GANCS goes beyond the MR reconstruction, and tailors other image restoration tasks dealing with aliasing artifacts. There are still important question to address such as using 3D spatial correlations for improved quality imaging, robustifying against patients with abnormalities, and variations in the acquisition model for instance as a result of different sampling strategies.


%


\newpage

\newpage



\medskip

\end{document}